\documentclass{ecai} 
\usepackage{latexsym}
\usepackage{amssymb}
\usepackage{amsmath}
\usepackage{amsthm}
\usepackage{booktabs}
\usepackage{enumitem}
\usepackage{graphicx}
\usepackage{color}
\usepackage[justification=centering]{caption}%标题名全体居中
\usepackage{subcaption}
\usepackage{algorithm}
\usepackage{algpseudocode} 
\usepackage{algorithmicx}

\newtheorem{definition}{Definition}

\newcommand{\BibTeX}{B\kern-.05em{\sc i\kern-.025em b}\kern-.08em\TeX}

\begin{document}

%%%%%%%%%%%%%%%%%%%%%%%%%%%%%%%%%%%%%%%%%%%%%%%%%%%%%%%%%%%%%%%%%%%%%%%%

\begin{frontmatter}

%%% Use this command to specify your submission number.
%%% In doubleblind mode, it will be printed on the first page.

\paperid{300} 

%%% Use this command to specify the title of your paper.

\title{$\mathrm{E^{2}CFD}$: Towards Effective and Efficient Cost Function Design
for Safe Reinforcement Learning via Large Language Model}

%%% Use this combinations of commands to specify all authors of your 
%%% paper. Use \fnms{} and \snm{} to indicate everyone's first names 
%%% and surname. This will help the publisher with indexing the 
%%% proceedings. Please use a reasonable approximation in case your 
%%% name does not neatly split into "first names" and "surname".
%%% Specifying your ORCID digital identifier is optional. 
%%% Use the \thanks{} command to indicate one or more corresponding 
%%% authors and their email address(es). If so desired, you can specify
%%% author contributions using the \footnote{} command.

% \author[A]{\fnms{First}~\snm{Author}\orcid{....-....-....-....}\thanks{Corresponding Author. Email: somename@university.edu.}\footnote{Equal contribution.}}
% \author[B]{\fnms{Second}~\snm{Author}\orcid{....-....-....-....}\footnotemark}
% \author[B,C]{\fnms{Third}~\snm{Author}\orcid{....-....-....-....}} 

% \address[A]{Short Affiliation of First Author}
% \address[B]{Short Affiliation of Second Author and Third Author}
% \address[C]{Short Alternate Affiliation of Third Author}

\author[A]{\fnms{Zepeng}~\snm{Wang}\footnote{Equal contribution.}}
\author[A]{\fnms{Chao}~\snm{Ma}\footnote{Equal contribution.}}
\author[A]{\fnms{Linjiang}~\snm{Zhou}} 
\author[A]{\fnms{Libing}~\snm{Wu}} 
\author[B]{\fnms{Lei}~\snm{Yang}} 
\author[A]{\fnms{Xiaochuan}~\snm{Shi}\thanks{Corresponding Author. Email: shixiaochuan@whu.edu.cn.}}
\author[A]{\fnms{Guojun}~\snm{Peng}\thanks{Corresponding Author. Email: guojpeng@whu.edu.cn.}} 

\address[A]{Key Laboratory of Aerospace Information Security and Trusted Computing, Ministry of Education, School of Cyber Science and Engineering, Wuhan University, China}
\address[B]{School of Software Engineering, South China University of Technology, China}

%%% Use this environment to include an abstract of your paper.

\begin{abstract}
Different classes of safe reinforcement learning algorithms have shown satisfactory performance in various types of safety requirement scenarios. However, the existing methods mainly address one or several classes of specific safety requirement scenario problems and cannot be applied to arbitrary safety requirement scenarios. In addition, the optimization objectives of existing reinforcement learning algorithms are misaligned with the task requirements. Based on the need to address these issues, we propose $\mathrm{E^{2}CFD}$, an effective and efficient cost function design framework. $\mathrm{E^{2}CFD}$ leverages the capabilities of a large language model (LLM) to comprehend various safety scenarios and generate corresponding cost functions. It incorporates the \textit{fast performance evaluation (FPE)} method to facilitate rapid and iterative updates to the generated cost function. Through this iterative process, $\mathrm{E^{2}CFD}$ aims to obtain the most suitable cost function for policy training, tailored to the specific tasks within the safety scenario. Experiments have proven that the performance of policies trained using this framework is superior to traditional safe reinforcement learning algorithms and policies trained with carefully designed cost functions.
\end{abstract}

\end{frontmatter}

%%%%%%%%%%%%%%%%%%%%%%%%%%%%%%%%%%%%%%%%%%%%%%%%%%%%%%%%%%%%%%%%%%%%%%%%

\section{Introduction}

Currently, safety requirements play a vital role in various fields. Different safety requirement scenarios cover a wide range of application fields, including autonomous driving \cite{isele2018safe}, recommendation systems \cite{ge2021towards}, resource allocation \cite{bhatia2019resource}, etc. In order to cope with these complex safety requirements, safe reinforcement learning algorithms have gradually become an effective solution in recent years. They take task requirements and safety requirements as optimization objectives for learning and training, and finally obtain policies that meet the safety requirements of corresponding scenarios. However, existing safe reinforcement learning algorithms still have some problems:

%     \begin{figure}[h]
%     \setlength{\abovecaptionskip}{0.2cm}
%     \setlength{\belowcaptionskip}{0.2cm}
% 	% \centering
% 	\begin{subfigure}{0.5\linewidth}
% 		\centering
% 		\includegraphics[width=0.99\linewidth]{Figures/Figure1/reward.pdf}
% 		\caption{reward}
% 	\end{subfigure}
%         % \centering
%         \hfill
% 	\begin{subfigure}{0.5\linewidth}
% 		\centering
% 		\includegraphics[width=0.99\linewidth]{Figures/Figure1/cost.pdf}
% 		\caption{cost}
% 	\end{subfigure}
%  % \vspace
%  % \vspace
% 	\caption{Performance between the human-engineered functions and original function.}
%  \label{fig:1}
%     \end{figure}
% \vspace

\begin{itemize}
    % \item \textbf{Poor generalization of safety reinforcement learning algorithms}: Most existing safe RL algorithms only design algorithms for a single or a small number of safety constraint scenarios, but there are many different safety requirements in real-world scenarios (such as cumulative constraint violation limits, zero constraint violation limits, almost surely satisfy constraint, etc.), a safety algorithm cannot be adapted to various safety requirement scenarios (such as simply setting the limit of cost in the algorithm that satisfies the cumulative constraint violation limit to 0, the final effect still cannot satisfy zero constraint violation). So how to design a unified safe RL algorithm framework that can be adapted to any safety constraint problem and ensure the effectiveness of the algorithm while improving the generalization of the algorithm has become a thorny issue.
    \item \textbf{Poor generalization of safe reinforcement learning algorithms.} Most existing safe reinforcement learning algorithms only design algorithms for a single or a small number of safety constraint scenarios, but there are many different safety requirements in real scenarios (e.g., cumulative constraint violation limit, zero-constraint violation limit, almost 100\% satisfaction of constraints, etc.). Traditional safety algorithms cannot adapt to various safety requirement scenarios by simply changing the algorithm parameter settings (e.g. just setting the cost limit to 0 in a safety algorithm that satisfies the cumulative constraint violation restriction still fails to satisfy the zero constraint violation \cite{he2023autocost}). Therefore how to design a generalized safe reinforcement learning algorithm framework that can adapt to any safety constraint problem, guarantee the effectiveness of the algorithm, and at the same time improve the generalization of the algorithm becomes a thorny issue.
 %    \begin{figure}[t]
	% % \centering
	% % \begin{subfigure}{0.48\linewidth}
	% 	\centering
	% 	\includegraphics[width=0.99\linewidth]{Figures/Figure1/performance.pdf}
	% 	\caption{PPO human designed reward}
	% % \end{subfigure}
 %    \end{figure}

    \begin{figure}[t]
    \setlength{\abovecaptionskip}{0.2cm}
    \setlength{\belowcaptionskip}{0.2cm}
	% \centering
	\begin{subfigure}{0.5\linewidth}
		\centering
		\includegraphics[width=0.99\linewidth]{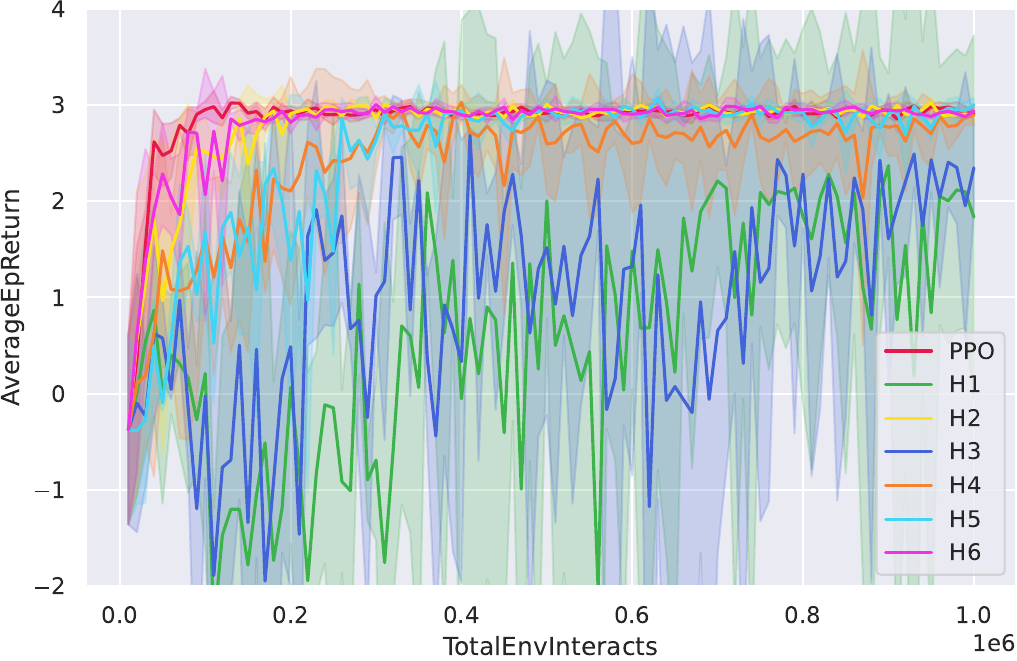}
		\caption{reward}
	\end{subfigure}
        % \centering
        \hfill
	\begin{subfigure}{0.5\linewidth}
		\centering
		\includegraphics[width=0.99\linewidth]{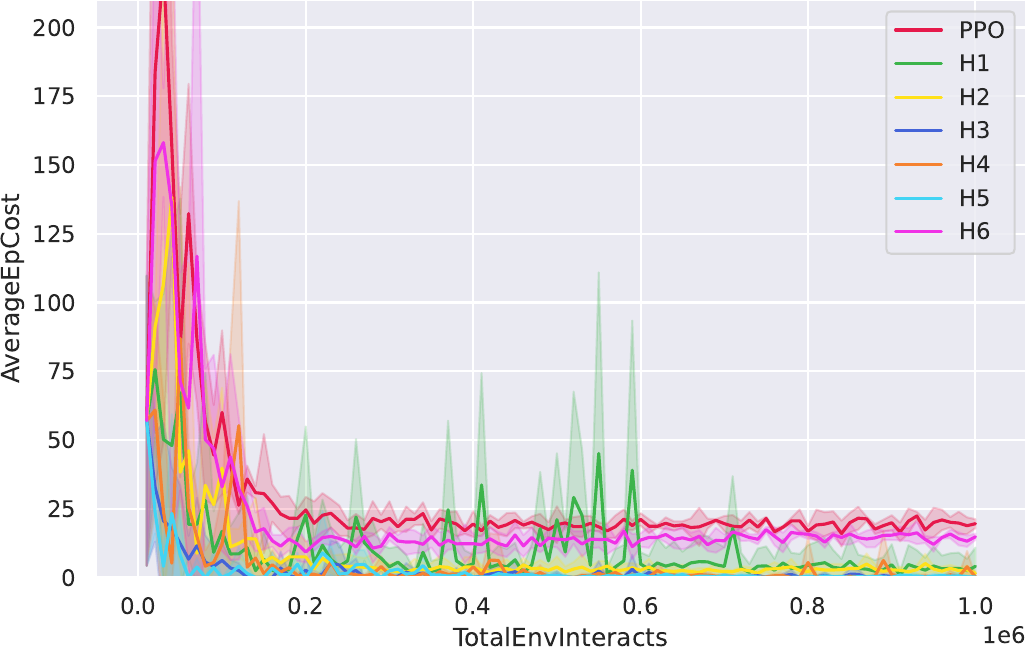}
		\caption{cost}
	\end{subfigure}
 % \vspace
 % \vspace
	\caption{Performance between the human-designed functions and original function.}
 \label{fig1}
    \end{figure}
% \vspace
    % \item \textbf{Alignment problem between RL algorithm optimization goal and task goal}: For any task, reinforcement learning as an artificial intelligence method, the goal is to hope that the agent can achieve the task (reach the goal), and the specific RL algorithm can only focus on improving the final cumulative reward value, which results in the quality of reward design being positively related to the task goal, but not completely consistent (for example, there is a reward hacking problem for this phenomenon). For a For complex task scenarios and goals, it is impossible to design a perfect reward so that the final optimization goal (maximizing cumulative reward) is equal to the goal of the task. For safe RL scenarios, the final performance evaluation indicators cannot only focus on the performance of reward and cost, but also focus on the actual task completion status (whether the task is completed and whether safety requirements are violated). Therefore, different reward/cost functions have different performance effects on the algorithm. As shown in Figure 1, it is a graph showing the difference in performance of different manually designed cost functions for the PPO algorithm. This inspires us to realize different safety task requirements by designing appropriate cost functions.

    \item \textbf{Alignment problem between RL algorithm optimization goal and task goal.} The goal of reinforcement learning as an AI approach is to have agents that achieve actual task goals. The structure of the RL algorithm, on the other hand, dictates that its optimization goal focuses on increasing the final cumulative reward value, which results in an algorithm whose optimization goal is positively correlated with the task goal, but not perfectly aligned (e.g., reward hacking problem \cite{skalse2022defining}). For a complex task scenario and objective, it is difficult to design the perfect reward function such that the final optimization objective of the RL algorithm is completely equivalent to the actual task objective. For a safe reinforcement learning scenario, the final evaluation metrics can likewise not only focus on the performance of rewards and costs, but also on the actual completion of the task (whether the task is completed or not, and whether the safety requirements are violated or not). We noticed that different qualities of reward and cost functions have different impacts on the performance of the algorithm (Figure~\ref{fig1} reflects the performance difference between using different human-designed reward functions in the PPO algorithm). This inspires us to realize different safety task requirements by designing appropriate reward functions and cost functions.
    
    % \item \textbf{The difficult problem of reward design}: For manual reward shaping, it requires a lot of time and professionals to try. Similar to parameter adjustment, the time and labor costs are large, and rewards need to be redesigned for different scenarios; for using LLM to do rewards The generated work is currently not used in complex scenarios with safety constraints, and most of them make use of LLM's ability to decompose tasks rather than directly generating rewards.

    \item \textbf{The difficult design of reward function.} Designing reward functions manually requires a lot of time and expertise. Similar to parameter tuning, the time and labor costs are greater. In dealing with more complex scenarios where safety constraints exist, there are even more factors to consider. When there is a slight change in the environment or task requirements, the designed reward function often fails and needs to be redesigned again according to the new environment and task requirements. Therefore, the manual design of reward functions remains problematic in the safety problem scenarios that are the focus of this paper.
\end{itemize}

% On the other hand, LLM has performed well in many fields recently. LLM's task understanding and code generation capabilities are also helpful in training many reinforcement learning tasks. However, most of the existing LLM-assisted RL training methods are based on task decomposition (robot-related tasks), and for some task planning that does not require complex tasks, LLM plays a limited role. How to use the advantages of LLM to assist reinforcement learning algorithm training in safety scenarios has become another problem that needs to be solved.

On the other hand, LLM has recently excelled in areas such as robot control \cite{yu2023language,huang2023inner}. The task understanding and code generation capabilities of LLM are useful for training many reinforcement learning tasks \cite{cao2024survey}. However, most of the existing LLM-assisted RL training methods consider goal decomposition for multi-step task operations, and LLM plays a limited role in scenarios that do not require task planning for complex tasks. How to utilize the advantages of LLM to assist in the training of reinforcement learning algorithms in safety scenarios is currently an open problem.

% In order to solve the above-mentioned problems, we designed a new secure reinforcement learning algorithm framework. The framework first uses a large language model to understand the tasks of different safety scenarios, initializes and generates different cost functions according to specific safety requirements to replace the original cost function of the environment, and evaluates the newly generated cost functions through the \textit{Faster Performance Evaluation } method used for rapid performance evaluation of reinforcement learning policy training, the evaluation results are used to generate a new cost function, and the iterative update process is repeated to finally obtain an optimal cost function and optimal policy for the safety requirement scenario.

% To address the above problems, we design a new framework, $\mathrm{E^{2}CFD}$, for cost function design solutions that can also be used as a new framework for safety reinforcement learning algorithms. We first define the cost function design problem. To solve this problem, we propose a cost function generation method that utilizes the comprehensive task understanding and code generation capabilities of LLM. In addition, we introduce an \textit{error code filtering (ECF)} module to ensure the effectiveness of code generation, as well as a specially tailored \textit{faster policy evaluation (FPE)} module to facilitate efficient and humane policy generation. The experimental results confirm the efficiency and effectiveness of our proposed framework. 
Our solution design idea is to fully utilize the advantages of LLM to solve the above problems. Specifically, we can use the task comprehension ability of LLM to solve the generalizability problem of safe reinforcement learning algorithms, adopt a new evaluation approach to solve the alignment problem of task requirements and optimization objectives, and use the code generation ability of LLM to solve the difficult problem of reward function design.
Overall, our contributions are as follows:
\begin{itemize}
    \item We first formulate the \textbf{Cost Definition Problem} \textit{(CDP)} and transform the task of solving different safety requirements into the task of designing different cost functions.
    % \item We propose a new safe reinforcement learning algorithm framework, which uses large language models to generate different cost functions for different task requirements and conducts rapid performance evaluation. It can be applied to any reinforcement learning algorithm to achieve policy training.
    \item We propose a new cost function generation framework that utilizes the comprehensive task understanding and code generation capabilities of LLM. In addition, we introduce an \textbf{Error Code Filtering} \textit{(ECF)} module to ensure the effectiveness of code generation, as well as a specially tailored \textbf{Fast Performance Evaluation} \textit{(FPE)} module to facilitate efficient and user-friendly policy generation.
    \item We conduct extensive and detailed experiments on a continuous control task. The experimental results show that our proposed framework provides better performance, generalization, and interpretability than traditional safe reinforcement learning algorithms with artificial reward function design methods.
\end{itemize}

%%%%%%%%%%%%%%%%%%%%%%%%%%%%%%%%%%%%%%%%%%%%%%%%%%%%%%%%%%%%%%%%%%%%%%%%

\section{Related Work}

In recent years, safe RL has derived various methods to solve problems with different levels of safety requirements. The most common safety requirement is the constraint that the cumulative constraints should be less than a certain threshold. Currently, most safe RL methods are designed based on this requirement. Commonly used methods include Lagrangian-based method \cite{ding2020natural,stooke2020responsive,paternain2022safe}, Projected-based method \cite{yang2019projection}, Lyapunov-based method \cite{chow2018lyapunov}, Safeguard-based method \cite{dalal2018safe}, etc. 
% The second safety requirement is to hope that a certain constraint will be satisfied almost 100\% (i.e., satisfy constraints almost surely), which puts forward higher requirements for the stability of the safety constraint guarantee of the algorithm. The current solution to this requirement is mainly based on state augmentation \cite{sootla2022saute,wang2023ccpo}. 
The second safety requirement is for the agent to achieve zero violation of constraints \cite{zhao2021model}. Meeting this requirement often requires the imposition of hard safety constraints on each step of the agent's behavior, which often leads to a decline in task target performance and an increase in training difficulty. 
% The most stringent safety requirement is for the agent to achieve zero violation of constraints \cite{zhao2021model}. Meeting this requirement often requires the imposition of hard safety constraints on each step of the agent's behavior, which often leads to a decline in task target performance and an increase in training difficulty. 
The most stringent safety requirement is to hope that a certain constraint (or zero violation) will be satisfied almost 100\% (i.e., satisfy constraints almost surely), which puts forward higher requirements for the stability of the safety constraint guarantee of the algorithm. The current solution to this requirement is mainly based on state augmentation \cite{sootla2022saute,wang2023ccpo}. 
How to design a reinforcement learning algorithm that can satisfy various safety requirement levels is still a current research challenge. 

Designing an appropriate reward function for a reinforcement learning agent to achieve the desired goal is difficult \cite{dewey2014reinforcement}. AutoRL \cite{faust2019evolving} proposes to use an evolutionary algorithm approach to automatically search for better reward functions. LIRPG \cite{zheng2018learning} proposes the idea of learning intrinsic rewards. AutoCost \cite{he2023autocost} proposes to use evolutionary strategies to automatically design cost functions to solve zero constraint violation problems.

Some language-conditioned-based RL works \cite{carta2022eager,mirchandani2021ella} use large models to understand the original task and then decompose it into subtasks to aid agent learning. LEARN \cite{goyal2019using} trains a classification network on trajectory data with natural language, to predict whether a trajectory matches the linguistic description, and evaluate the network at each time step to form a potential-based reward shaping function. Recently, due to the exciting performance of LLM in task understanding and code generation, some works have used LLM to help reinforcement learning task training. Colas et al. \cite{colas2020language} automatically generates goals from textual descriptions of the tasks and uses them for subsequent training of the agent. Some works directly model LLM as a reward function. Kwon et al. \cite{kwon2022reward} uses LLM to generate binary reward functions and Eureka \cite{ma2023eureka} proposes a framework for reward function generation for robot control tasks using LLM. However, there is currently little work in the field of safe reinforcement learning that uses LLM to aid in training.

To the best of our knowledge, we are the first work to use LLM for cost function design in the safe RL domain. Specifically, the most similar works to ours are AutoCost \cite{he2023autocost} and Eureka \cite{ma2023eureka}. However, the former does not take advantage of LLM to understand the task and generate code automatically, while the latter uses the complete environment source code as additional information input to better understand the task and generate code, and the problem scenarios do not involve more complex safety requirement constraints. In addition, both use more inefficient evolutionary algorithms that are not suitable for direct application to safe RL problems. Instead, our proposed $\mathrm{E^{2}CFD}$ only requires privileged access to the necessary information (environment description, task description, form of the original reward function and cost function) as auxiliary information. We believe that in a gray-box scenario (i.e., where part of the environment-related information can be accessed to assist the LLM in task comprehension, in addition to agent's observations and feedback), the less information used to assist training, the better the generalizability of the method.

    \begin{figure*}[h]
    \setlength{\abovecaptionskip}{0.2cm}
    \setlength{\belowcaptionskip}{0.2cm}
	% \centering
	% \begin{subfigure}{0.48\linewidth}
		\centering
            \includegraphics[width=0.99\linewidth]{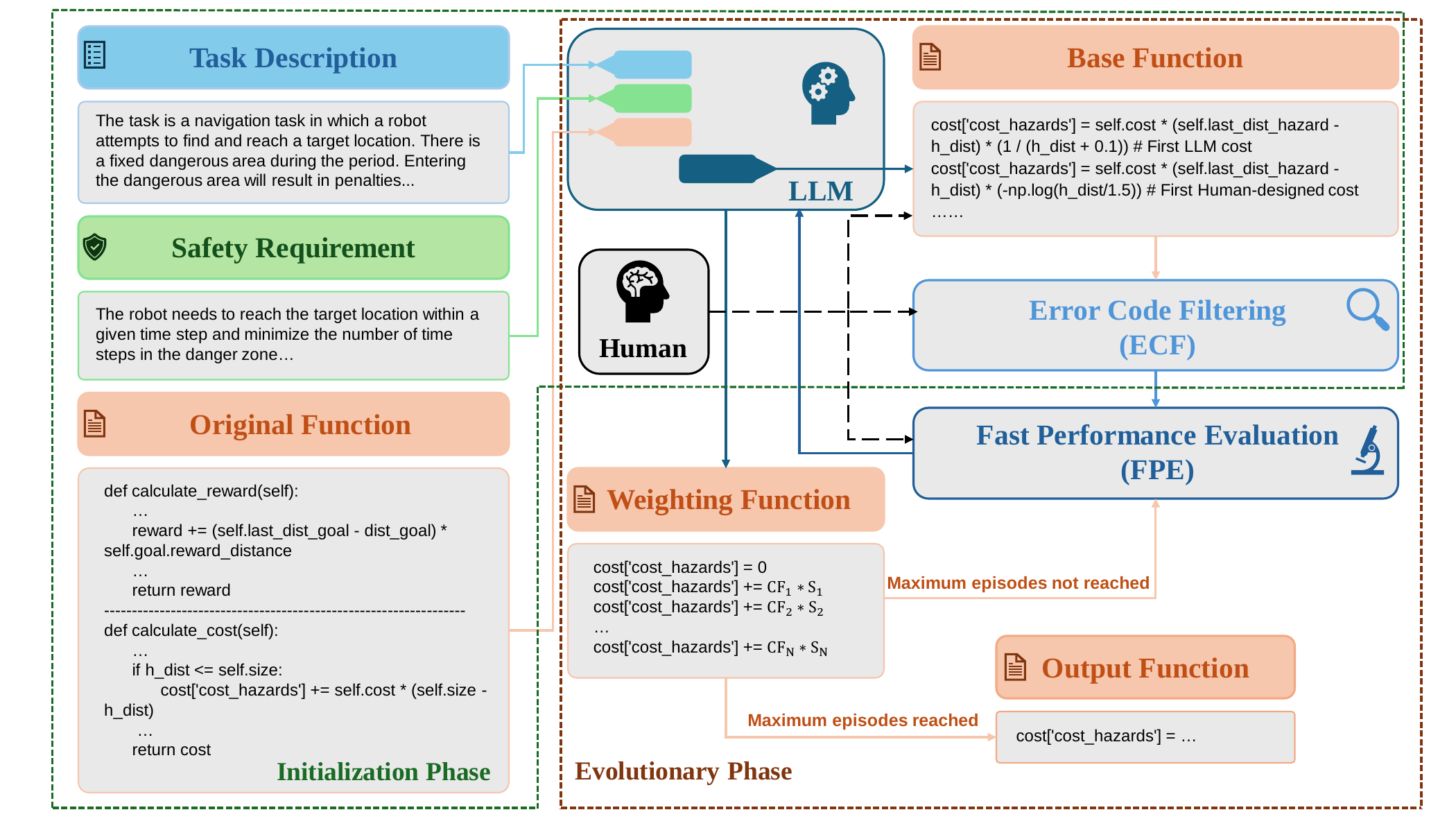}
		\caption{Framework of $\mathrm{E^{2}CFD}$.}
	% \end{subfigure}
 \label{fig2}
    \end{figure*}

%%%%%%%%%%%%%%%%%%%%%%%%%%%%%%%%%%%%%%%%%%%%%%%%%%%%%%%%%%%%%%%%%%%%%%%%

\section{Background and Problem Formulation}

\subsection{Constrained Markov Decision Process}
The constrained Markov decision process (CMDP) is usually used as a modeling method for safe reinforcement learning. It is usually defined as a tuple $(S, A, P, R, C, \mu, \gamma)$ consisting of the following parts: the state space $S$, the action space $A$, the transition probability function $P: S\times A\times S\rightarrow[0,1]$ describing the dynamics of the environment, the reward function $R: S\times A\rightarrow \mathbb{R}$ representing the immediate reward obtained by taking action $a$ in state $s$, the cost function $C: S\times A\rightarrow \mathbb{R}$ representing the immediate cost of taking action $a$ in state $s$, the initial state distribution $\mu: S\rightarrow[0,1]$ and the discount factor $\gamma\in[0,1]$ determining the emphasis on future gains.

In safe RL, the optimization goal is to obtain a policy $\pi$ that maximizes the discounted cumulative reward $J_{R}^{\pi}={\mathbb E}_{\tau\sim\pi}\left[\sum_{t=0}^{\infty}\gamma^{t}R(s_t,a_t)\right]$ and at the same time, the discounted cumulative cost $J_{C}^{\pi}={\mathbb E}_{\tau\sim\pi}\left[\sum_{t=0}^{\infty}\gamma^{t}C(s_t,a_t)\right]$ satisfies specific safety constraints, expressed as:

\begin{equation}
\begin{aligned}
\max \limits_{\pi\in\Pi} \enspace J_{R}^{\pi},\quad {s.t.}\enspace SF(\pi) 
\end{aligned} \label{eq1}
\end{equation}

\noindent where $SF(\pi)$ denotes different safety requirements. Take three common safety constraint requirements as an example:
\begin{itemize}

\item The \textit{traditional safety requirement} requires that the discounted cumulative cost is satisfied to be less than a certain deterministic threshold $d$, then $SF(\pi): J_C^{\pi}-d \leq 0$.

\item The \textit{zero-violation safety requirement} requires that the discounted cumulative cost is 0, then $SF(\pi): J_C^{\pi} \leq 0$.

\item The more complex \textit{almost surely safety requirement} requires close to 100\% satisfaction of a constraint, then $SF(\pi): z_t \geq 0\enspace a.s., \forall t \geq 0$, where $z_t$ indicates whether the discounted cumulative cost has violated the safety constraint when in the current state $s_t$ and $a.s.$ stands for "almost surely" (with probability one).

\end{itemize}

It is worth mentioning that regarding the way different safety constraints are defined, our approach is more general and concise compared to other existing work \cite{yao2024constraint}, which is suitable for extension to modeling various safety scenarios or other scenarios with constraints.

\subsection{Cost Design Problem}
% Since the task objectives of real scenarios (e.g., the success rate of the task and the number of risky behaviors, etc.) are often difficult to use directly for optimization using reinforcement learning algorithms as a reward function and a cost function, these metrics are generally only statistically available at the end of training and separately in the testing phase. The goal of cost function design is therefore to design a suitable cost function that is aligned with the real task goal and is able to achieve the task goal by using the newly designed cost function. To this end, referring to the traditional way of defining the reward design problem \cite{singh2009rewards}, we first propose a new definition of the cost function design problem.

The task objectives of real scenarios are often statistical quantities (e.g., the success rate of the task and the number of dangerous behaviors, etc.), which can usually only be counted individually at the end of training and during the testing phase, and which cannot be directly optimized as a reward function and a cost function directly using a reinforcement learning algorithm. Therefore, the goal of cost function design is to design a suitable cost function that matches the actual task objectives and to be able to utilize the newly designed cost function to achieve the task objectives. To this end, referring to the traditional way of defining the reward design problem \cite{singh2009rewards}, we first propose a new definition of the cost design problem.

\begin{definition}[Cost Design Problem]
    A \textbf{cost design problem} (CDP) is a tuple $P=\langle M,\pi_M,F \rangle$, where M is a CMDP. $\pi_M$ is a policy obtained by training based on reward function $R$ and cost function $C$ using any learning algorithm. $F:\pi\rightarrow\mathbb{R}$ is the fitness score function for generating a scalar evaluation result score for an arbitrary policy, which can only be obtained through the actual policy evaluation process. In a CDP, the goal is to output a cost function $C$ so that the trained policy achieves the highest score $F(\pi)$.
\end{definition}

Ideally, the fitness score function should be aligned with the ultimate goal of the task. For example, for unconstrained scenarios with no safety requirements, the metric degenerates into the cumulative reward under the standard reinforcement learning task setting, i.e., $F(\pi) = J_R^{\pi}$. While for traditional safety-constrained scenarios where safety requirements exist, the metric can be expressed as:

\begin{equation}
F(\pi)=\left\{
\begin{aligned}
    & -n & J_C^{\pi}\textgreater d, \\
    & J_R^{\pi}  & J_C^{\pi}\leq d. 
\end{aligned}
\right.
\label{eq2}
\end{equation}

\noindent where $d$ is the safety constraint, $n$ is a sufficiently large positive number, $J_R^{\pi}$ is the discount cumulative reward, and $J_C^{\pi}$ is the discount cumulative cost.

%%%%%%%%%%%%%%%%%%%%%%%%%%%%%%%%%%%%%%%%%%%%%%%%%%%%%%%%%%%%%%%%%%%%%%%%

\section{Methodology}

The schematic of our proposed framework is shown in Figure~\ref{fig2} and is divided into an initialization phase and an evolutionary phase.

 %    \begin{figure*}[h]
 %    \setlength{\abovecaptionskip}{0.2cm}
 %    \setlength{\belowcaptionskip}{0.2cm}
	% % \centering
	% % \begin{subfigure}{0.48\linewidth}
	% 	\centering
	% 	% \includegraphics[width=0.99\linewidth, height=12cm]{Figures/ECF.pdf}
 %            \includegraphics[width=0.99\linewidth]{Figures/Framework6.pdf}
	% 	\caption{Framework of $\mathrm{E^{2}CFD}$.}
	% % \end{subfigure}
 % \label{fig2}
 %    \end{figure*}

\subsection{Initialization Phase}

% \subsubsection{Step 1: Base Function Generation}
\subsubsection{Base Function Generation}
The first step in the initialization phase is to perform base cost function generation. Depending on the specific environment description and safety task requirements, there can be two types of generation: 1. Large language model for generating the cost function; 2. Manual design for generating the cost function. Our framework jointly uses these two approaches to collaboratively generate base cost functions to obtain diverse and efficient initial base cost functions.

% Although it was previously mentioned that performing cost function design manually is both time-consuming and laborious, designing only an imperfect cost function is not costly and helps in initialization.
% Another approach is to utilize the task comprehension capability of the large language model to generate an initial cost function by combining task description information, safety requirement information, and original reward function and cost function templates already in the environment. The advantage of this approach is that it is automated, convenient, and does not require manual time to understand and analyze the task.
% Our framework uses both approaches for base cost function generation to obtain a diversity of base cost functions for initialization.
%Our framework jointly uses these two approaches to collaboratively generate base cost functions to obtain diverse and efficient initial base cost functions. %The framework utilizes the task comprehension capability of the large language model to generate a series of initial cost functions by combining task description information, safety requirement information, and the original reward function and cost function templates already available in the environment. 

First of all, we need to input the necessary information related to the task as prompts to LLM. For task scenarios with different task requirements, the necessary information contains:
\begin{enumerate}
    \item \textbf{Task description information.} This information passes the semantic content of the task (including environment-base information, task objectives, etc.) into the LLM for subsequent selection of appropriate elements as components of the cost function.
    \item \textbf{Safety requirement information.} This information passes the task's safety requirement content (including the degree of safety constraints, safety objectives, etc.) into the LLM for the generation of safety components in the cost function.
    \item \textbf{Original reward/cost function information.} This information passes the original reward/cost function content (including code style, function variables, etc.) into the LLM for the generation of code fragments in the cost function to guarantee syntactic accuracy and semantic consistency.
\end{enumerate}
Meanwhile, although it was mentioned earlier that manually designing cost functions is both time-consuming and labor-intensive, it is not costly to design only imperfect cost functions. Therefore the framework combines a series of suboptimal cost functions obtained without sufficient manual design with the cost functions generated by the large language model to obtain the final set of initial cost functions, which helps to ensure a lower bound on the quality of the initialized functions.

% \subsubsection{Step 2: Error Code Filtering}
\subsubsection{Error Code Filtering}
\begin{figure}[h]
    \setlength{\abovecaptionskip}{0.2cm}
    \setlength{\belowcaptionskip}{0.2cm}
	% \centering
	% \begin{subfigure}{0.48\linewidth}
		\centering
            \includegraphics[width=0.99\linewidth]{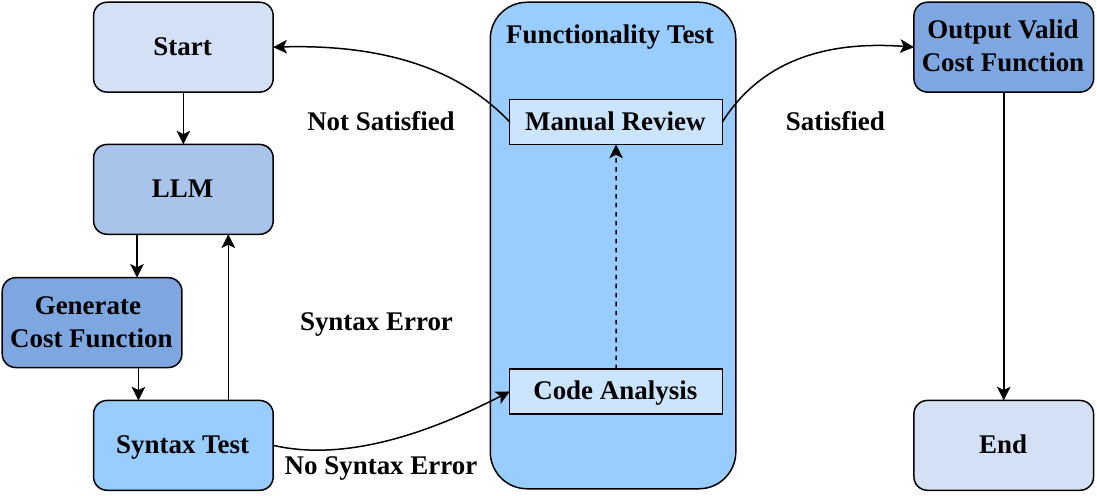}
		\caption{Structual diagram of \textbf{Error Code Filtering} \textit{(ECF)} module.}
	% \end{subfigure}
\label{fig3}
    \end{figure}

Due to the imperfections in the human-designed prompts, the human intent cannot be fully conveyed to the LLM, and thus the output of the LLM often contains some errors (non-compliance, obvious syntax errors, etc.). We therefore propose the following \textbf{Error Code Filtering} \textit{(ECF)} module to ensure that the cost function code used for subsequent training is free of syntax errors and satisfies human intent as much as possible. Figure~\ref{fig3} illustrates the specific construction of the \textit{ECF} module.

% Specifically, the \textit{ECF} module first performs a syntax test on the generated cost function. By simply replacing the generated cost function with the previous one, a complete round of training process is run, and if no compilation error (syntax error) occurs during the process, the newly generated cost function is considered to be free of syntax error.
% The \textit{ECF} then eliminates new cost functions that clearly do not meet the mission requirements (e.g., the presence of components that contradict the safety requirements) by introducing a manual review. The final output of syntax error-free cost functions that potentially satisfy the mission requirements is used in the next phase.

Specifically, the \textit{ECF} module first performs a syntax test on the generated cost function. If no compilation errors (syntax errors) occur during this process, the newly generated cost function is considered free of syntax errors.
Then, \textit{ECF} rejects new cost functions that clearly do not meet the task requirements (e.g., the presence of components that contradict safety requirements) by introducing manual review. The final output of cost functions that are free of syntax errors and likely to fulfill the task requirements will be used in the next phase.

\subsection{Evolutionary Phase}

% \subsubsection{Step 1: Fast Performance Evaluation}
\subsubsection{Fast Performance Evaluation}
In the evolutionary phase, we use an evolutionary algorithm-like approach to evaluate the base function generated by iterative updating to obtain a better cost function to aid in the training of the agent.

It is worth stating that after obtaining the generated cost function, it is an open question of how to use the cost function for agent training. Similar to the treatment of many classical safe reinforcement learning algorithms (e.g., Lagrangian-based methods), we chose to put the cost function directly into the reward function as a penalizing term in the reward function, and later use the modified reward function as a new reward function for training using any reinforcement learning algorithm. 

\begin{equation}
R(s_t,a_t)=R(s_t,a_t)+C(s_t,a_t) \label{eq3}
\end{equation}

% This seemingly simple treatment has the following benefits: first, this enables training using any standard reinforcement learning algorithm, and second, it further simplifies the cost of manually designing the function by transferring the intractable problem of weighting between the reward function and the cost function, as in the traditional approach, to a part of the \textit{CFDP}.

This seemingly simple treatment makes it possible to train using any standard reinforcement learning algorithm. At the same time, it further simplifies the cost of manually designing the function by transferring the weighting problem between the reward function and the cost function, which is difficult to solve in traditional methods, to a part of \textit{CDP}.

However, according to the traditional evolutionary algorithm approach, it is very inefficient to wait until the agent is trained (e.g., curve convergence) before evaluating its performance in a test environment alone. Therefore, we propose a new policy performance evaluation method, \textbf{Fast Performance Evaluation} \textit{(FPE)}.

First, policy training of the agent is started using one weighting cost function or multiple base cost functions generated in the previous step. Then, when a predefined number of early training rounds is reached, training is stopped and performance evaluation is performed based on a predefined scoring function. Finally, the performance evaluation results are used to iteratively update the weighting cost function or base cost functions.

% In this case, the scoring function (similar to the fitness function in genetic algorithms) can either be set manually and predefined or generated by LLM scoring. To reduce the cost of manual design, our framework uses LLM to generate the scoring function. We compare the effects of different scoring functions on the performance results in the experimental section.

Furthermore, as mentioned before, the fitness score function that perfectly aligns with the final optimization goal of the task can only be obtained by evaluating it in a separate test at the end of agent training. Scenarios with different safety level requirements, on the other hand, need to be manually crafted to define different fitness functions that are consistent with the task requirements. Therefore, the $\mathrm{E^{2}CFD}$ framework abandons the practice of manually defining the fitness score function, and instead uses the LLM to generate appropriate fitness score functions based on the task requirements, in order to reduce the design cost and increase the flexibility of the framework to face different scenarios. We compare the effects of different fitness score functions on the performance results in the experimental section.

Overall the \textit{FPE} module has two advantages. First, it improves the efficiency of policy performance evaluation and can help to obtain a cost function that helps to accelerate the convergence of the training of the agent. Second, it solves the problem of aligning the task objective with the reinforcement learning training objective.

% \subsubsection{Step 2: Weighting Function Generation}
\subsubsection{Weighting Function Generation}
Using the performance evaluation results from the first step, a new weighting cost function can be generated by using the output of the fitness score function as the importance weight of each cost function. Specifically, for the set of $n$ base cost functions $F^{b}=\{f^{b}_1, f^{b}_2, ..., f^{b}_n\}$ generated in the first step, which corresponds to the fitness scores obtained as $S=\{s_1, s_2,...,s_n\}$, the newly generated weighting function is:

\begin{equation}
f^{w}=\sum_{i=1}^n f^{b}_i \cdot s_i \label{eq4}
\end{equation}
% Note that we did not normalize the scores, and we found that using the LLM scores directly as weights for subsequent training without going through normalization worked satisfactorily. We believe it is that the LLM scoring takes into account both the weights between the reward function and the cost function, as well as the weights of the different cost functions, and therefore no further normalization is required to obtain the appropriate weights.
where $S$ is the fitness score after the normalization process, as it is important to avoid the imbalance between the cost function and the original reward function caused by too high or too low LLM scores.

% If the new weighting cost function still does not satisfy the task requirement, then we can continue to use LLM again for new base function generation based on this newly obtained weighting cost function, repeating the previous process (refer to 4.1 and 4.2) to achieve iterative updating until the task requirement is satisfied or the maximum number of iteration rounds is reached.

The subsequent iterative updating and optimization process of the cost function can be achieved by simply using the newly generated cost function as an input to the LLM in the initialization phase and repeating the previous process until the maximum number of iteration rounds is reached.

\subsection{Overall Algorithm Framework}
The pseudo-code of the complete algorithmic framework is shown in Algorithm~\ref{alg1}.

\begin{algorithm}
    \caption{$\mathrm{E^{2}CFD}$}
    \begin{algorithmic}[1] % The number tells where the line numbering should start
        % \State \textbf{Input:} Task Description $TD$, Safety Requirement $SR$, Original Function $OF$, Large Language Model $LLM$, Update iterations $N$, Base Function Number $K$, Early Evaluation Stage $t_1$, Late Evaluation Stage $t_2$.
        \Require
            Task description $TD$, safety requirement $SR$, original function $f^{o}$, large language model $LLM$, update iterations $N$, the number of base cost functions $K$, early evaluation phase $t_1$, and late evaluation phase $t_2$.
        \State \textbf{/* Base Function Generation */}
        \State $f^{w}_{best} = \text{None}$ \hfill\textbf{//  best weighting function}
        % \State $f^{s} = LLM(TD, SR)$ \hfill\textbf{//  fitness score function}
        \State $p_{best} = -\infty$ \hfill \textbf{//  best performance score}
        \State $F^{b} = LLM.function\_generation(TD, SR, f^{o}, f^{w}_{best})$
        % \State $WF_{best} = \text{None}$
        \For{$n = 1$ \textbf{to} $N$}
            \State \textbf{/* Error Code Filtering */}
            % \State $f^{b}_1, \ldots, f^{b}_K = ECF(f^{b}_1, \ldots, f^{b}_K)$
            \State $F^{b} = ECF(F^{b})$
            \State \textbf{/* Fast Performance Evaluation */}
            % \State $p_1, \ldots, p_K = FPE(f^{b}_1, \ldots, f^{b}_K, t_1)$
            \State $p_1, \ldots, p_K = FPE(F^{b}, t_1)$
            % \State $s_1, \ldots, s_K \gets LLM(TD, SR, BF_1, \ldots, BF_K, p_1, \ldots, p_K)$
            \State \textbf{/* Weighting Function Generation */}
            \State $S = LLM.score\_evaluation(TD, SR, F^{b}, p_1, \ldots, p_K, f^{s})$
            \State $f^{w}=\sum_{i=1}^K f^{b}_i \cdot s_i$
            \State $p_{tmp} = FPE(f^{w}, t_2)$
            % \State $p' = FPE(WF, t_2)$
            \If{$p_{tmp} > p_{best}$}
                \State $f^{w}_{best} = f^{w}$
                \State $p_{best} = p_{tmp}$
                % \State $p_{best} = p'$
            \EndIf
            \If{$n < N$}
                \State \textbf{/* Base Function Generation */}
                \State $F^{b} = LLM.function\_generation(TD, SR, f^{o}, f^{w}_{best})$
            \EndIf
        \EndFor
        % \State \textbf{Output:} $WF_{\text{best}}$
        \Ensure
            Best cost function $f^{w}_{best}$.
    \end{algorithmic}
    \label{alg1}
\end{algorithm}

%%%%%%%%%%%%%%%%%%%%%%%%%%%%%%%%%%%%%%%%%%%%%%%%%%%%%%%%%%%%%%%%%%%%%%%%

\section{Experiments}

\subsection{Experiment Settings}
We evaluate the proposed framework $\mathrm{E^{2}CFD}$ on the StaticPointGoal task in the static environment version of Safety Gym \cite{yang2021wcsac,ray2019benchmarking}, a benchmark widely used for safe RL algorithm evaluation. The task exists of a Point robot with 46 states and 2 actions navigating towards a goal in a 2D space while avoiding contact with hazardous areas while traveling. The initial position of the robot is randomly generated and the current episode ends when the robot reaches the goal. Note that the locations of the hazardous regions of the environment are fixed, which we do in order to carry out a more intuitive interpretation of the subsequent construction process of the cost function, and to avoid unsolvable problems contaminating our experiments \cite{yang2021wcsac,sootla2022saute}.

All agents were trained for 100 epochs with 10,000 interaction steps per epoch and a maximum step size of 1,000 steps per episode. All experiments are run using three random seeds. 
% We choose six algorithms (one unconstrained algorithm and five safe RL algorithms), PPO, PPO-Lag, CPO, PCPO, FOCOPS, and CUP, as the comparison baselines algorithms, and the code base is based on SPO (NIPS2023). 
For a fair comparison, we compare our $\mathrm{E^{2}CFD}$ with five recognized classical or state-of-the-art safe reinforcement learning algorithms (PPO-Lag, CPO \cite{achiam2017constrained}, PCPO \cite{yang2019projection}, FOCOPS \cite{zhang2020first}, and CUP \cite{yang2022constrained}) under the same codebase \cite{ji2023safety}. We also use the standard PPO \cite{schulman2017proximal} algorithm as a representative of unconstrained algorithms in all subsequent experiments to refer to the normal task performance versus the safety performance in this task scenario. All algorithms adopt the same training strategy and tricks except for the extra components of the algorithm itself.
The complete code is also placed in a supplemental file and will be open-sourced when it is subsequently organized well.
% The specific experimental setup and parameters are described in Appendix.

In addition to using cumulative returns and cumulative costs as traditional task metrics and safety metrics, we also used three metrics, task completion rate (TCR), hazardous area exposure rate (HER), and time ratio (TR), in some of our experiments as more comprehensive metrics to reflect the effectiveness and efficiency performance of our algorithmic framework in aligning human needs. The specific definitions of the three metrics are as follows:

\begin{equation}
\left\{
\begin{aligned}
    &TCR = \frac{n_{tc}}{n_e} \\
    &HER = \frac{n_{hae}}{n_e} \\
    &TR = \frac{t_{algo}}{t_{ppo}}
\end{aligned}
\right.\label{eq5}
\end{equation}

\noindent where $n_{tc}$ is the number of task completion, $n_{hae}$ is the number of hazardous area exposure, $n_e$ is the number of episode, $t_{algo}$ is the training time of algorithm and $t_{ppo}$ is the training time of PPO.

\subsection{Performance under Different Safety Requirements}
\subsubsection{Traditional Safety Requirement Scenarios}
    
    \begin{figure}[htbp]
    \setlength{\abovecaptionskip}{0.2cm}
    \setlength{\belowcaptionskip}{0.2cm}
	\centering
	\begin{subfigure}{0.48\linewidth}
		\centering
		\includegraphics[width=0.99\linewidth]{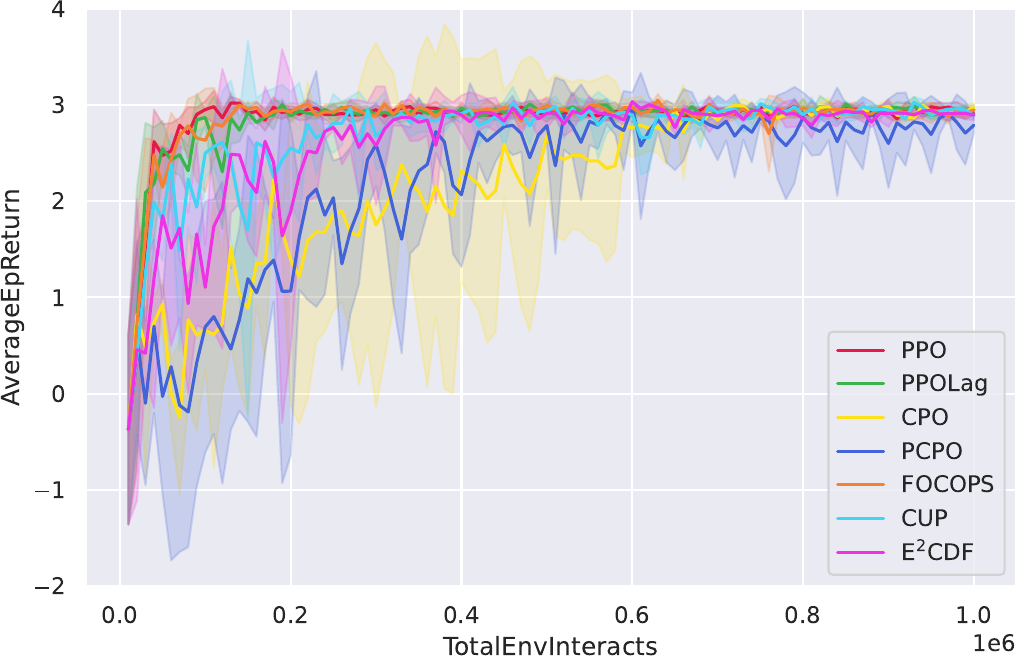}
		\caption{reward}\label{fig:sub1}
	\end{subfigure}
        \centering
	\begin{subfigure}{0.48\linewidth}
		\centering
		\includegraphics[width=0.99\linewidth]{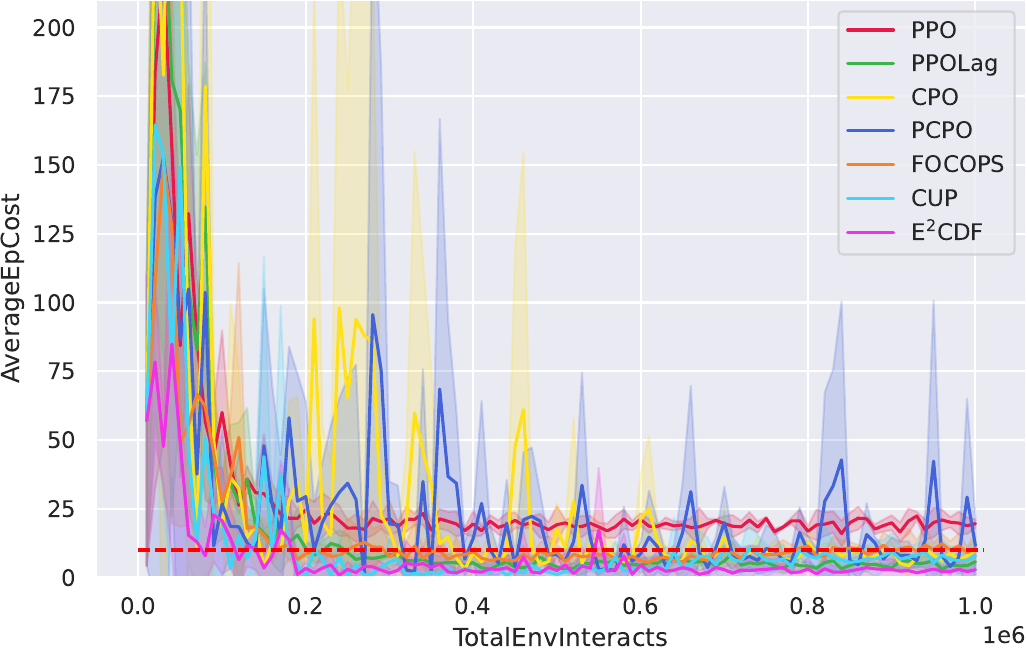}
		\caption{cost}\label{fig:sub2}
	\end{subfigure}
 %        \centering
 %        \begin{subfigure}{0.99\linewidth}
	% 	\centering
	% 	\includegraphics[width=0.99\linewidth]{Figures/Figure2/cost_box.pdf}
	% 	\caption{boxplot of cost}\label{fig:sub3}
	% \end{subfigure}
	\caption{Performance on traditional safety requirement scenarios. In this scenario, for the traditional safe RL algorithm, we set $d=10$ for the expected cost limit (black dashed line).}
 \label{fig4}
    \end{figure}

 We first compare the performance of the proposed $\mathrm{E^{2}CFD}$ with the baseline algorithms on traditional safety requirement tasks. Figure~\ref{fig4} shows that all algorithms eventually meet the task requirements and all algorithms except PPO and PCPO meet the safety requirements. However, we find that $\mathrm{E^{2}CFD}$ significantly outperforms all other algorithms in terms of safety performance, which reflects the potential of $\mathrm{E^{2}CFD}$ in safety requirement fulfillment.
 
 % In addition, by comparing the results of the box plots of costs in Figure~\ref{fig:sub3}, it can be observed that the overall distribution of constraint satisfaction for $\mathrm{E^{2}CFD}$ performs the best.

\subsubsection{Zero-violation Safety Requirement Scenarios}

    \begin{figure}[htbp]
    \setlength{\abovecaptionskip}{0.2cm}
    \setlength{\belowcaptionskip}{0.2cm}
	\centering
	\begin{subfigure}{0.48\linewidth}
		\centering
		\includegraphics[width=0.99\linewidth]{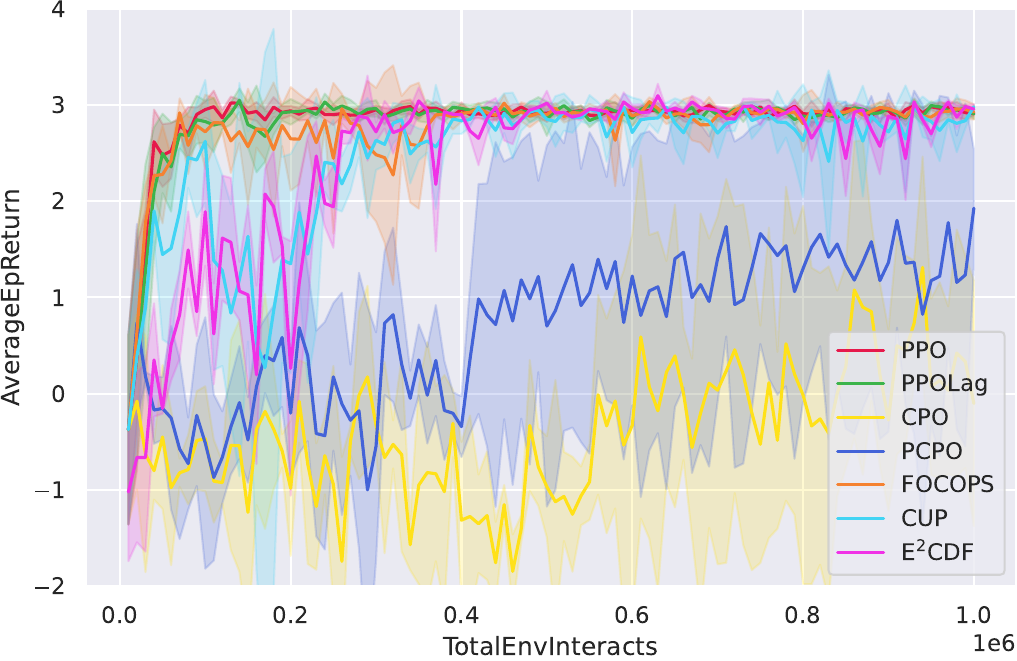}
		\caption{reward}\label{fig:sub4}
	\end{subfigure}
        \centering
	\begin{subfigure}{0.48\linewidth}
		\centering
		\includegraphics[width=0.99\linewidth]{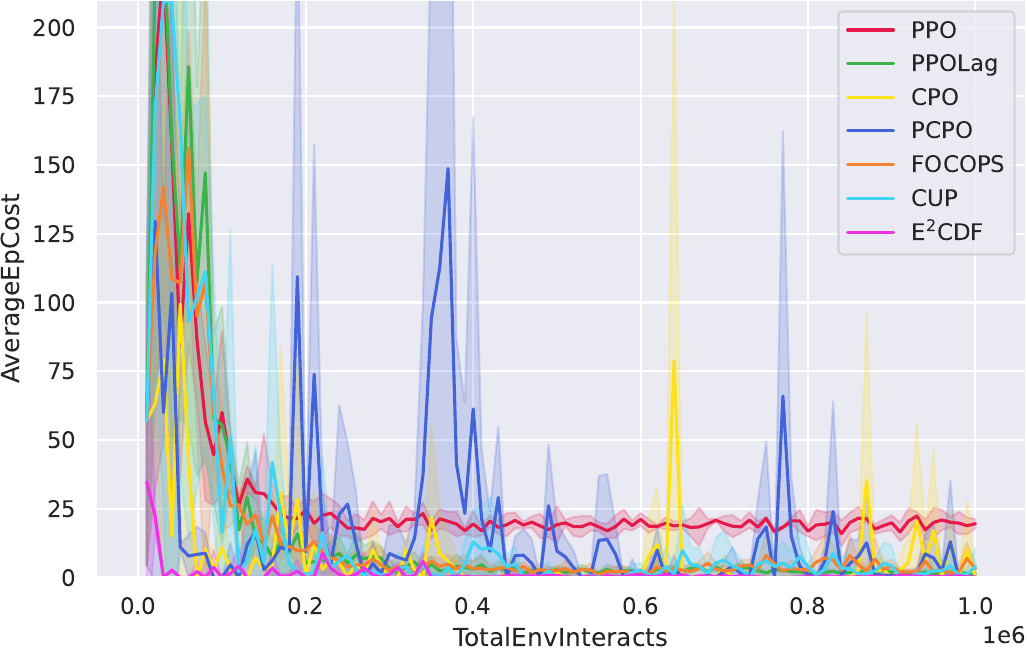}
		\caption{cost}\label{fig:sub5}
	\end{subfigure}
 %        \centering
 %        \begin{subfigure}{0.99\linewidth}
	% 	\centering
	% 	\includegraphics[width=0.99\linewidth]{Figures/Figure3/cost_box.pdf}
	% 	\caption{boxplot of cost}\label{fig:sub6}
	% \end{subfigure}
	\caption{Performance on zero-violation safety requirement scenarios. In this scenario, for the zero-violation safe RL algorithm, we set $d=0$ for the expected cost limit.}
        \label{fig5}
    \end{figure}

We also compare the performance of the proposed $\mathrm{E^{2}CFD}$ with the benchmark algorithms on the safety requirement task with zero violation constraints. Figure~\ref{fig5} shows that not all algorithms eventually meet the task requirements, which reflects the impact of more demanding safety requirement scenarios on the algorithms' task performance. $\mathrm{E^{2}CFD}$ is also the closest algorithm to zero-constraint violation in terms of safety requirements. This shows the obvious advantage of $\mathrm{E^{2}CFD}$ over traditional safe reinforcement learning algorithms in different safety requirement scenarios.

% Similarly, by comparing the box plot results of the costs in Figure~\ref{fig:sub6}, it can be observed that the $\mathrm{E^{2}CFD}$ performs second only to the CPO algorithm in terms of the overall distribution of constraint satisfactions, while significantly outperforming all the other baseline algorithms. However, it is worth emphasizing that the CPO algorithm performs very poorly on the original task with zero constraint violation of the safety requirements, while $\mathrm{E^{2}CFD}$F excels in both task performance and safety requirement performance.

\subsubsection{Almost surely Safety Requirement Scenarios}

This safety requirement scenario is an enhanced version of the first two safety scenarios. It emphasizes more on the stability of the algorithm's safety performance, i.e., it requires that the safety constraints (traditional safety constraints given a pre-threshold or zero violation safety constraints) are satisfied in as many episodes as possible.

\begin{figure}[htbp]
    \setlength{\abovecaptionskip}{0.2cm}
    \setlength{\belowcaptionskip}{0.2cm}

        \centering
        \begin{subfigure}{0.48\linewidth}
		\centering
		\includegraphics[width=0.99\linewidth]{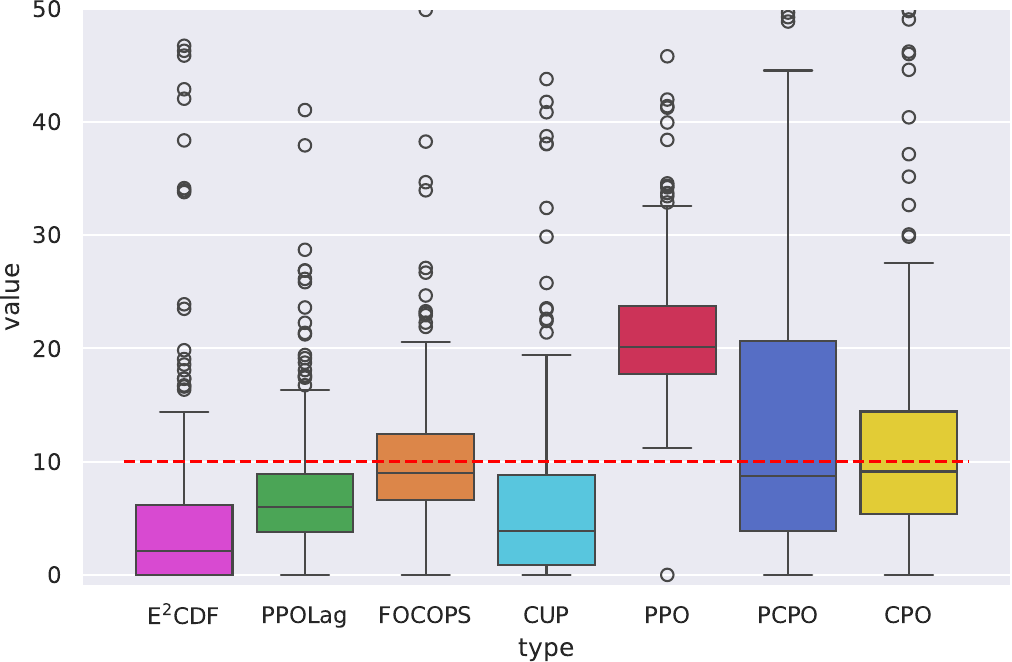}
		\caption{almost surely traditional safety requirement}\label{fig:sub3}
	\end{subfigure}
         \centering
        \begin{subfigure}{0.48\linewidth}
		\centering
		\includegraphics[width=0.99\linewidth]{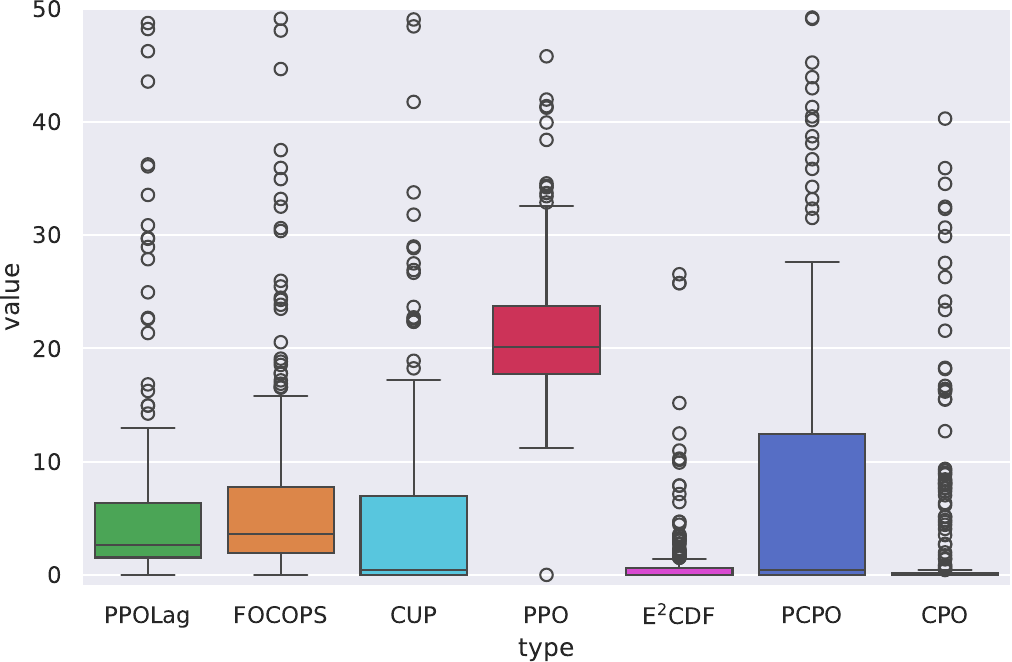}
		\caption{almost surely zero-violation safety requirement}\label{fig:sub6}
	\end{subfigure}
	\caption{Performance on almost surely safety requirement scenarios. Box plots show the overall satisfaction of the constraints and the distribution characteristics such as median, 75th and 25th quantiles and outlier points of the experimental results, which are helpful for evaluating whether the constraints are satisfied almost surely.}
 \label{fig6}
    \end{figure}

By comparing the results of the box plots of costs in Figure~\ref{fig:sub3}, it can be observed that the overall distribution of constraint satisfaction for $\mathrm{E^{2}CFD}$ performs the best. Similarly, by comparing the box plot results of the costs in Figure~\ref{fig:sub6}, it can be observed that the $\mathrm{E^{2}CFD}$ performs second only to the CPO algorithm in terms of the overall distribution of constraint satisfactions, while significantly outperforming all the other baseline algorithms. However, it is worth emphasizing that the CPO algorithm performs very poorly on the original task with zero constraint violation of the safety requirements (see Figure~\ref{fig:sub4}), while $\mathrm{E^{2}CFD}$F excels in both task performance and safety requirement performance.

\subsection{Human-engineered vs. LLM-assisted CFD}

\begin{figure}[htbp]
    \setlength{\abovecaptionskip}{0.2cm}
    \setlength{\belowcaptionskip}{0.2cm}
	\centering
	\begin{subfigure}{0.48\linewidth}
		\centering
		\includegraphics[width=0.99\linewidth]{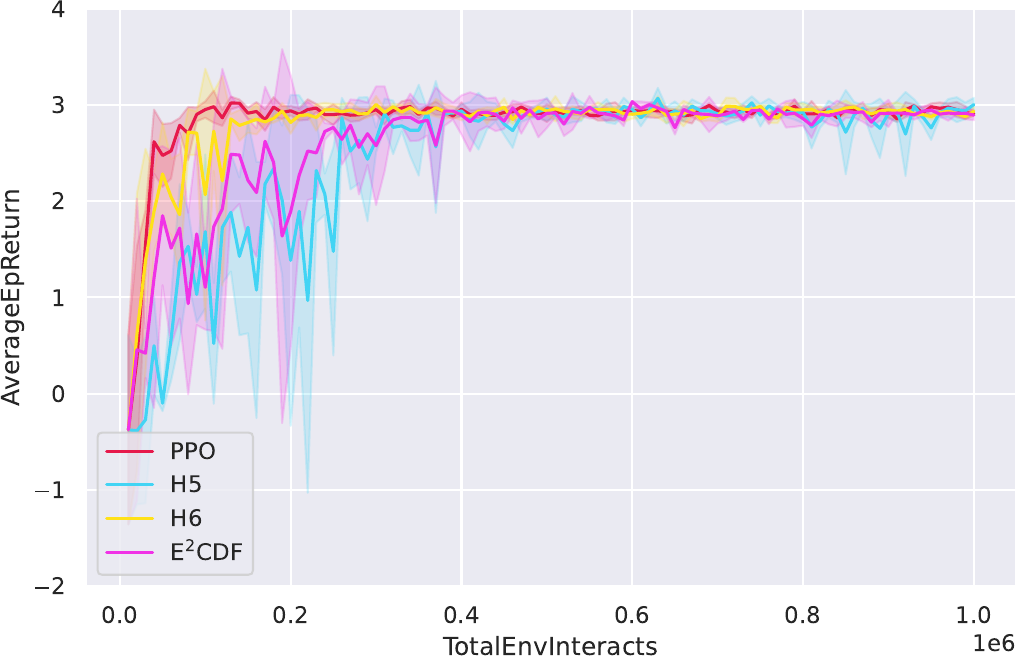}
		\caption{reward}
	\end{subfigure}
        \centering
	\begin{subfigure}{0.48\linewidth}
		\centering
		\includegraphics[width=0.99\linewidth]{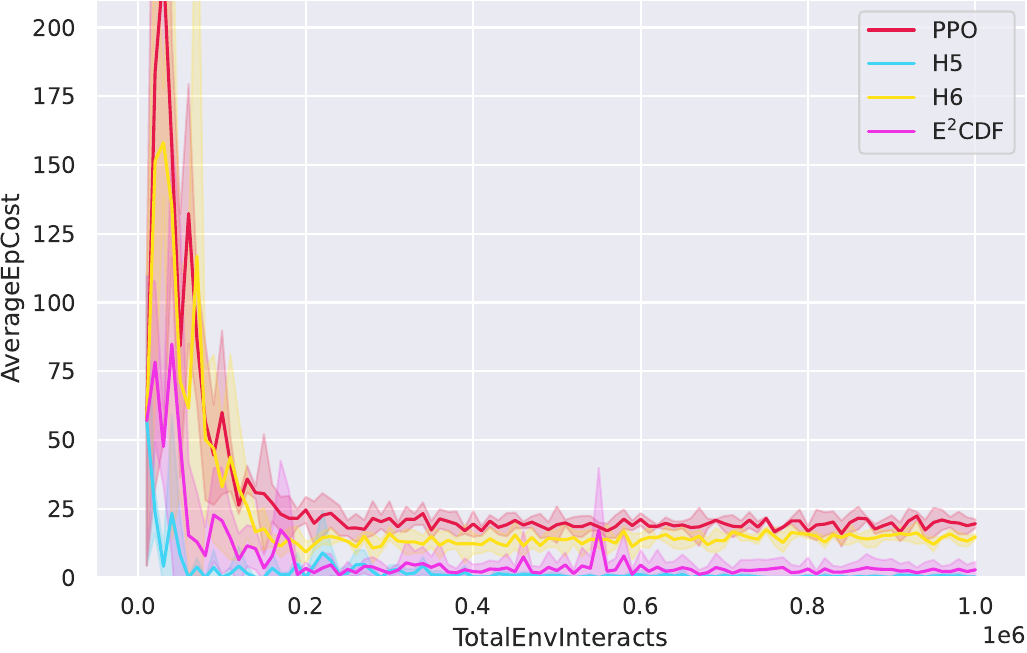}
		\caption{cost}
	\end{subfigure}
	\caption{Performance between the human-engineered cost functions and LLM-assisted cost functions.}
        \label{fig7}
    \end{figure}

Figure~\ref{fig7} compares the difference in task performance between the human-engineered cost functions and LLM-assisted cost functions. In this case, $\mathrm{H5}$ and $\mathrm{H6}$ are the two best methods in terms of safety performance and task performance, respectively, among all human-designed cost functions (consistent with the experimental results in Figure~\ref{fig1}).

The experimental results show that our proposed $\mathrm{E^{2}CFD}$ demonstrates superior competitiveness in terms of final performance compared to the performance of the method with a cost function that has been carefully designed manually, with the advantage that it does not need to spend a lot of cost for tuning parameter trial and error.

\subsection{FPE Model Evaluation}
% Different stages of evaluation / Different scoring criteria
In order to specifically analyze the \textit{FPE} module in our proposed framework $\mathrm{E^{2}CFD}$, we conducted comparative experiments on two of its key components (the evaluation phase and the scoring function).

    \begin{figure}[htbp]
    \setlength{\abovecaptionskip}{0.2cm}
    \setlength{\belowcaptionskip}{0.2cm}
	\centering
	\begin{subfigure}{0.49\linewidth}
		\centering
		\includegraphics[width=0.99\linewidth]{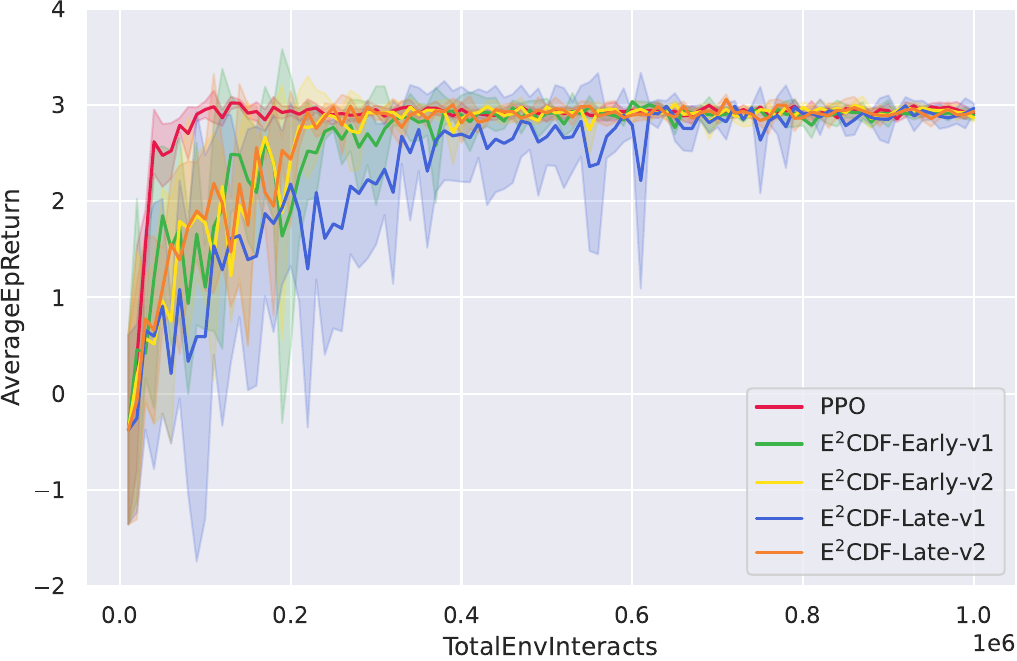}
		\caption{reward}
	\end{subfigure}
        \centering
	\begin{subfigure}{0.49\linewidth}
		\centering
		\includegraphics[width=0.99\linewidth]{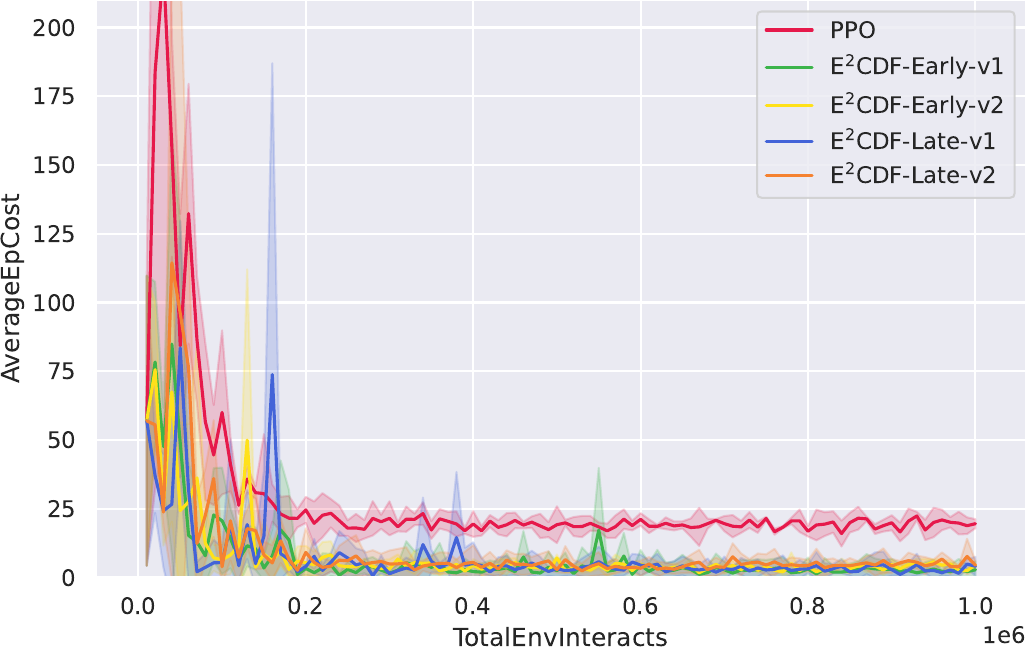}
		\caption{cost}
	\end{subfigure}
	\caption{Performance at different evaluation stages and with different scoring functions.}
 \label{fig8}
    \end{figure}

\begin{table}[htbp]
\setlength{\abovecaptionskip}{0.2cm}
\setlength{\belowcaptionskip}{0.2cm}
\begin{center}
{\caption{Performance at different evaluation phases and with different scoring functions. Bolding indicates the first two (at least) best-performing results. LLM-output1 and LLM-output2 denote two different scoring functions generated by LLM.}\label{table1}}
    \renewcommand{\arraystretch}{1} 
    \begin{tabular}{l@{\hspace{0.1cm}}c@{\hspace{0.1cm}}c@{\hspace{0.1cm}}c@{\hspace{0.1cm}}c@{\hspace{0.1cm}}c}
        % \hline
        \toprule 
        % \rule{0pt}{8pt}
        Algorithm      &      Evaluation Phase      &      Scoring Function      &      TCR ($\uparrow$)        &      HER ($\downarrow$)   &      TR ($\downarrow$)       \\
        % \hline
        % \rule{0pt}{8pt}
        \midrule 
        $\mathrm{PPO}$ & Late & N/A & \textbf{1.000} & 0.220 & \textbf{1.000}   \\ \cline{1-3}
        $\mathrm{E^{2}CFD}$ & Late & LLM-output1 & \textbf{1.000} &  \textbf{0.000} & 1.597  \\
        $\mathrm{E^{2}CFD}$ & Early & LLM-output1 & \textbf{1.000} & 0.023 & \textbf{1.084}   \\ \cline{2-3}
        $\mathrm{E^{2}CFD}$ & Late & LLM-output2 & 0.867 &  \textbf{0.013} & 1.792  \\
        $\mathrm{E^{2}CFD}$ & Early & LLM-output2 & \textbf{1.000} &  0.039 & 1.114  \\
        % \hline
        \bottomrule
    \end{tabular}
\end{center}
% \label{table1}
\end{table}

    \begin{figure*}[htbp]
    \setlength{\abovecaptionskip}{0.2cm}
    \setlength{\belowcaptionskip}{0.2cm}
	% \centering
	% \begin{subfigure}{0.48\linewidth}
		\centering
		\includegraphics[width=0.99\linewidth]{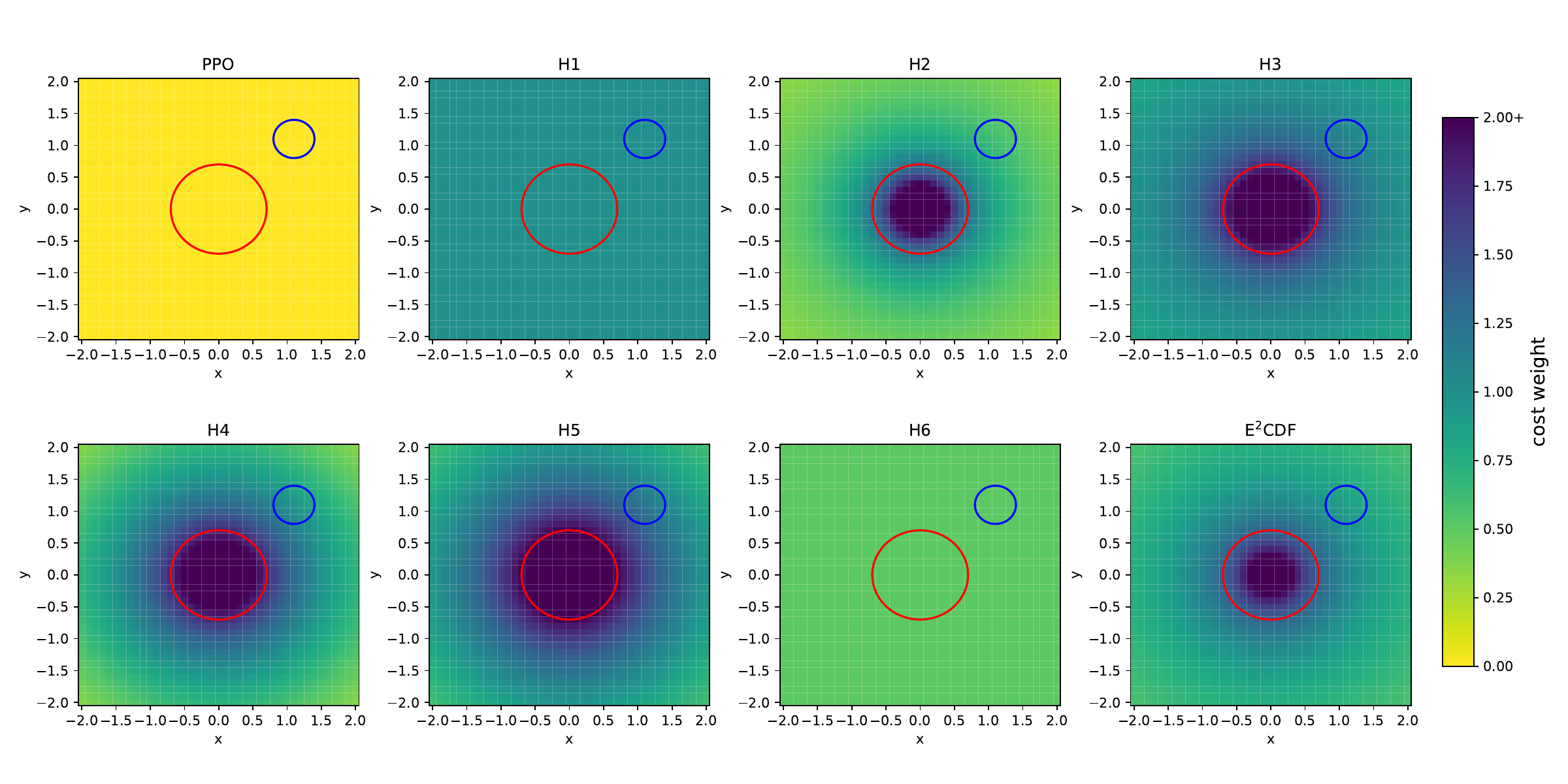}
		\caption{Visualization of cost function. In this case, the red circle indicates the range of the hazard and the blue circle indicates the range of the goal.}
	% \end{subfigure}
 \label{fig9}
    \end{figure*}

As can be obtained from Figure~\ref{fig8} and Table~\ref{table1}, whether we change the time node of performance evaluation, or replace the scoring function, eventually our framework $\mathrm{E^{2}CFD}$ demonstrates good dual optimization of task requirements (TCR) and safety requirements (HER). Specifically, for algorithms with different evaluation phases under the same scoring function, the algorithm that performs early performance evaluation shows a slight decrease in safety requirement fulfillment but a significant increase in time savings (TR) relative to the algorithm that performs late performance evaluation. For algorithms using different scoring functions (LLM-output1 or LLM-output2) under the same evaluation phase, the former corresponding algorithms show improved task performance, safety performance and time performance than the latter corresponding algorithms. The latter, on the other hand, exhibits a faster convergence rate, indicating its superior sample efficiency. This phenomenon reveals that we can improve the performance of the algorithmic framework by trying different scoring functions according to different needs.

\subsection{Visualization of Cost Functions}

To further analyze the cost functions generated by different approaches, we visualize the weight values in the composition of the cost functions in multiple algorithms (including PPO, human-designed, and $\mathrm{E^{2}CFD}$) to get the importance of the cost functions under different approaches for different regions of this static environment. 
% In this case, the red circle indicates the range of the hazard and the blue circle indicates the range of the goal.

The heatmap results in Figure~\ref{fig9} reflect the fact that the cost function obtained with the aid of the LLM has the same characteristics as the human-designed cost function, i.e., there is a perception of the hazardousness of the area in the environment. In addition, the level of safety emphasis in the vicinity of the goal area may also have an impact on the agent's policy.

The advantage of $\mathrm{E^{2}CFD}$ is that it does not need to rely entirely on manual coding of hazards in the environment, which provides a more convenient and efficient way of constructing cost functions for meeting other safety requirements in more complex environments, and has a greater potential for application. Meanwhile, in turn, the visualization and analysis of the cost function obtained by the LLM-assisted generation can help the human to better understand the task requirements, discover deeper design ideas, and also help the human to design a better cost function.

%%%%%%%%%%%%%%%%%%%%%%%%%%%%%%%%%%%%%%%%%%%%%%%%%%%%%%%%%%%%%%%%%%%%%%%%

\section{Conclusion}

In this paper, we present $\mathrm{E^{2}CFD}$, an effective and efficient cost function design framework for safe reinforcement learning via large language model. We first present the problem of generating cost functions under safety requirements for complex safety scenarios. Our approach leverages the task understanding and code generation capabilities of LLM and designs \textit{FPE} module to achieve efficient and human-aligned policy generation. The experiments demonstrate that the method has better performance in meeting task requirements than traditional safe reinforcement learning algorithms, and has the advantage of being more efficient and generalizable than the human-designed approach.

However, there are still directions for improvement in this framework. For example, LLM often fails to fully consider all the task requirements and precautions due to the incompleteness of the prompts, so the framework separately uses a manually-assisted \textit{ECF} module to screen the code reasonableness of generating base functions. How to design more complete prompts for task scenarios and improve the quality of LLM code generation will be a worthwhile research direction in the future. In addition, when the safety requirements increase (i.e., corresponding to multi-constraint scenarios), the complexity of the task will also increase. How to ensure the effectiveness of the framework to face such more complex problem scenarios will be another research difficulty.

\bibliography{mybibfile}

\end{document}